\documentclass[letterpaper, 10pt, conference]{IEEEtran}

\IEEEoverridecommandlockouts                              

\usepackage{xspace} 
\usepackage[T1]{fontenc} 
\usepackage{hyperref} 
\hypersetup{bookmarksopen,bookmarksnumbered,
pdfpagemode=UseOutlines,
colorlinks=true,
linkcolor=blue,
anchorcolor=blue,
citecolor=blue,
filecolor=blue,
menucolor=blue,
urlcolor=blue
}

\usepackage{url}  
\usepackage{graphicx}  
\usepackage{xspace} 
\usepackage{multicol} 
\usepackage{multirow} 
\usepackage{booktabs}
\usepackage{amsmath, amssymb, amsfonts,amsthm}
\usepackage{commath}
\usepackage{nicefrac}
\usepackage{dsfont}
\usepackage{microtype}
\usepackage{mathtools}
\usepackage{xfrac}
\usepackage{verbatim}
\usepackage{xcolor}
\usepackage{tikz}
\usepackage{algorithm}
\usepackage[noend]{algpseudocode}
\usepackage[font=footnotesize]{caption}
\usepackage[font=footnotesize]{subcaption}
\usepackage{enumitem}
\usepackage{accents}
\usepackage[export]{adjustbox}
\usepackage{color}
\usepackage[binary-units]{siunitx}
\usepackage[numbers]{natbib} 
\usepackage{cleveref}

\usepackage{pgfplots}
\pgfplotsset{compat=newest}
\pgfplotsset{every axis legend/.append style={%
cells={anchor=west}}
}
\usepgfplotslibrary{polar}
\usetikzlibrary{arrows}
\tikzset{>=stealth'}

\definecolor{gn_3_1}{RGB}{229,245,224}
\definecolor{gn_3_2}{RGB}{161,217,155}
\definecolor{gn_3_3}{RGB}{49,163,84}
\definecolor{or_3_1}{RGB}{254,230,206}
\definecolor{or_3_2}{RGB}{253,174,107}
\definecolor{or_3_3}{RGB}{230,85,13}
\definecolor{bu_3_1}{RGB}{222,235,247}
\definecolor{bu_3_2}{RGB}{158,202,225}
\definecolor{bu_3_3}{RGB}{49,130,189}

\newcommand{\calA}{\ensuremath{\mathcal{A}}\xspace}
\newcommand{\calS}{\ensuremath{\mathcal{S}}\xspace}
\newcommand{\calN}{\ensuremath{\mathcal{N}}\xspace}

\newcommand{\bfx}{\ensuremath{\mathbf{x}}\xspace}
\newcommand{\bfp}{\ensuremath{\mathbf{p}}\xspace}
\newcommand{\bfu}{\ensuremath{\mathbf{u}}\xspace}

\newcommand{\lmbd}{\ensuremath{\lambda_{d}}\xspace}
\newcommand{\lmbh}{\ensuremath{\lambda_{h}}\xspace}

\begin{document}
%
\title{
\vspace{0.25in} Dynamic Real-time Multimodal Routing \\ with Hierarchical Hybrid Planning
}

\author{\IEEEauthorblockN{Shushman Choudhury}\thanks{This research is funded through the Ford-Stanford Alliance.}
\IEEEauthorblockA{Department of Computer Science\\
Stanford University \\
\texttt{shushman@stanford.edu}}
\and
\IEEEauthorblockN{Jacob P. Knickerbocker}
\IEEEauthorblockA{Ford Greenfield Labs\\
Palo Alto \\
\texttt{jknicker@ford.com}}
\and
\IEEEauthorblockN{Mykel J. Kochenderfer}
\IEEEauthorblockA{Department of Aeronautics and Astronautics\\
Stanford University \\
\texttt{mykel@stanford.edu}}
}

\maketitle

\begin{abstract}
We introduce the problem of Dynamic Real-time Multimodal Routing (DREAMR), which requires planning and executing routes under uncertainty for an autonomous agent. The agent can use multiple modes of transportation in a dynamic transit vehicle network. For instance, a drone can either fly or ride on terrain vehicles for segments of their routes. DREAMR is a difficult problem of sequential decision making under uncertainty with both discrete and continuous variables. We design a novel hierarchical hybrid planning framework to solve the DREAMR problem that exploits its structural decomposability. Our framework consists of a global open-loop planning layer that invokes and monitors a local closed-loop execution layer. Additional abstractions allow efficient and seamless interleaving of planning and execution. We create a large-scale simulation for DREAMR problems, with each scenario having hundreds of transportation routes and thousands of connection points. Our algorithmic framework significantly outperforms a receding horizon control baseline, in terms of elapsed time to reach the destination and energy expended by the agent.
\end{abstract}

\section{Introduction}
\label{sec:intro}

Cost-effective transportation often requires multiple modes of movement, such as walking or public transport. Consequently, there has been extensive work on multimodal route planning~\cite{pajor2009multi,botea2013multi,delling2013computing}. Prior work, even state-of-the-art~\cite{huang2018multimodal,giannakopoulou2018multimodal}, has focused on generating static routes in public transit networks for humans, rather than real-time control for autonomous agents. This paper introduces a class of problems that we call Dynamic Real-time Multimodal Routing (DREAMR), where a controllable agent operates in a dynamic transit network.

\begin{figure}[t]
    \centering
    \fbox{\includegraphics[width=0.62\columnwidth]{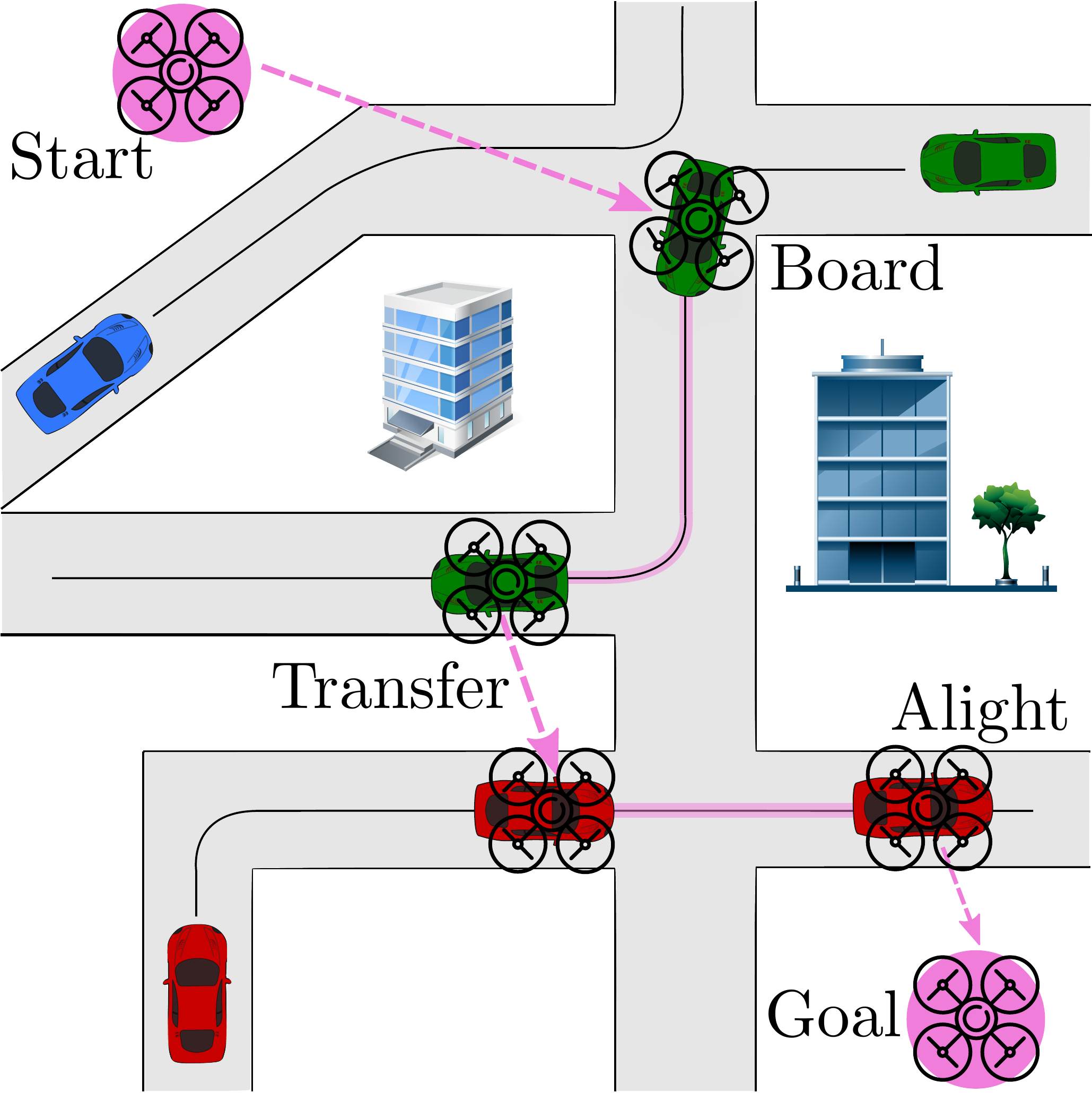}}
    \caption{Our planning framework makes real-time decisions for an agent to traverse a network of dynamic transit routes. The agent can ride on transit vehicles for segments of their own routes to save energy and enhance its effective range. The example of drones riding on cars is purely illustrative; our framework permits other agent dynamics and transit route properties.}
    \label{fig:fig1}
\end{figure}

An example of DREAMR is a drone riding on cars (\cref{fig:fig1}) en route to its destination. Such coordination can achieve energy efficiency and resilience for navigation, search-and-rescue missions, and terrain exploration~\cite{ponda2015cooperative}. Multimodal routing already helps humans schedule vehicle charging~\cite{schuster2012multimodal} and aerial-ground delivery systems have been developed~\cite{arbanas2016aerial}.

The DREAMR problem class inherits the challenges of multimodal route planning --- the combinatorial complexity of mode selection and the balancing of performance criteria like time and distance. In addition, the online transit network requires adaptivity to new information (new route options, delays and speedups). We need to make decisions at multiple time-scales by planning routes with various modes of transport, and executing them in real-time under uncertainty.


Our key insight is that DREAMR has significant underlying structure and decomposability to simpler sub-problems. 
We exploit this through a hierarchical hybrid planning framework. A global open-loop layer repeatedly plans (discrete) routes on the dynamic network through efficient graph search, deciding which modes of transport to use and for what duration. A local closed-loop layer executes (continuous) agent control actions in real-time under uncertainty. We design abstractions that interleave planning and execution efficiently to respond to updates in the transportation options.

Our contributions are as follows: (i) We introduce DREAMR and formulate it as an Online Stochastic Shortest Path problem~\cite{neu2010online} of both discrete and continuous variables. (ii) We propose a novel Hierarchical Hybrid Planning (HHP) framework for DREAMR. The choices of representation and abstraction in our framework adapt techniques for multimodal routing and hierarchical stochastic planning in an efficient yet principled manner. (iii) We create a synthetic setup for a subclass of DREAMR (single controllable agent; transportation assumptions) with hundreds of vehicles and significant route variability. Our framework shows scalability to very large problems and a better tradeoff between energy and elapsed time than a receding horizon control baseline.


\section{Background and Related Work}
\label{sec:background}

We discuss a background topic and two related work areas.

\subsection{Markov Decision Processes}
\label{sec:background-mdp}

A Markov Decision Process or MDP~\cite{puterman1994markov} is defined by $\left(\calS, \calA, T, R, \gamma\right)$, where \calS and \calA are the system's state and action spaces, $T\left(s,a,s^{\prime}\right) = P(s^{\prime} \mid s,a)$ is the transition function, where $s, s^{\prime} \in \calS$ and $a \in \calA$, $R(s,a)$ is the reward function, and $\gamma$ the discount factor. Solving an MDP yields a \emph{closed-loop} policy $\pi : \calS \times \calA \rightarrow [0,1]$ which maximizes the \emph{value} or the expected reward-to-go from each state. An MDP can be solved by \emph{value iteration}, a dynamic programming (DP) method that computes the optimal value function $V^{*}$.
We obtain a single $V^{*}$ for infinite horizon problems and $V^{*}_{0:K}$ ($K$ is the maximum number of timesteps) for finite horizon or episodic problems.
For large or continuous spaces, we can approximate the value function locally with multilinear interpolation~\cite{davies1996multidimensional}, or globally with basis functions~\cite{busoniu2010reinforcement}. Our framework uses approximate DP extensively~\cite{bertsekas2005dynamic}.

\subsection{Multimodal Route Planning}
\label{sec:background-mmrp}
 
Route planning in transportation networks has been studied considerably~\cite{bast2016route}. DREAMR involves multimodal route planning~\cite{pajor2009multi}, where multiple modes of transportation can be used. Multimodality can provide flexibility and resilience~\cite{bast2013result}, but introduces combinatorial and multiobjective complexity~\cite{modesti1998utility}. Heuristic methods have tried to address this complexity~\cite{delling2013computing}. Recent work has focused on real-world aspects like contingencies~\cite{botea2013multi,botea2015contingent}, dynamic timetable updates~\cite{giannakopoulou2018multimodal} and integrating more flexible carpooling~\cite{huang2018multimodal}.

\subsection{Hierarchical Stochastic Planning}
\label{sec:background-hierarchical}

A powerful technique for large MDPs is to decompose it into more tractable components~\cite{boutilier1999decision}. A hierarchical stochastic planning framework combines subproblem solutions to (approximately) solve the full MDP~\cite{dean1995decomposition}. We use various ideas from hierarchical reinforcement learning~\cite{barto2003recent}. Temporally abstract actions or \emph{macro-actions}~\cite{sutton1999between} are actions persisting over multiple timesteps and are viewed as local policies executed in a subset of $\calS$ until termination~\cite{hauskrecht1998hierarchical}. A hierarchy of abstract machines~\cite{parr1998hierarchical} is a framework with a multi-layer structure where higher-level supervisors monitor lower-level controllers and switch between them in appropriate parts of the state space. State abstraction improves tractability by ignoring aspects of the system that do not affect the current decision~\cite{andre2002state,dietterich2000hierarchical}.

\section{Problem Formulation}
\label{sec:problem}
We use a discrete-time undiscounted episodic MDP to model DREAMR.
The system state at time $t$ is
\begin{equation}
\label{eq:dreamr-state}
s_t = (\bfx_{t}, \bfp_{t}^{1}, \xi_{t}^{1}, \bfp_{t}^{2}, \xi_{t}^{2}, \ldots \bfp_{t}^{n_t}, \xi_{t}^{n_t}, I_{t})
\end{equation}
where $\bfx_{t}$ is the agent state, $\bfp_{t}^{i}$ is the current position of vehicle $i$, $\xi_{t}^{i}$ is the vehicle's remaining route, a sequence of waypoints with estimated time of arrival (ETA), i.e. $\xi_{t}^{i} = ((\bfp_{1,t}^{i},\tau_{1,t}^{i}),(\bfp_{2,t}^{i},\tau_{2,t}^{i}), \ldots)$, and $I_{t} \in \left\{0,1,\ldots,n_t\right\}$ indicates the vehicle the agent is riding on (0 if not). Due to the time-varying number of active vehicles $n_t$, the state space has \emph{nonstationary dimensionality}. The action space is the union of agent control space $\mathcal{U}$ and agent-vehicle interactions, i.e. $\calA \equiv \mathcal{U} \cup \left\{\textsc{Board}, \textsc{Alight} \right\}$. \emph{Both
state and action spaces are a hybrid of discrete and continuous variables}.

\textbf{Agent}: We use a 2D second-order point mass model for the agent (simplified to focus on decision-making, not physics). The state has position and bounded velocity and the control has bounded acceleration in each direction independently, subject to zero-mean Gaussian random noise,
\begin{equation}
\label{eq:drone-dynamics}
\begin{split}
&\bfx_{t} = \left(x_t,y_t,\dot{x}_t,\dot{y}_t\right), \ \bfu_t = \left(\ddot{x}_t,\ddot{y}_t\right)\\
&\bfx_{t+\delta t} = f(\bfx_t,\bfu_t + \epsilon,\delta t), \ \epsilon \sim \calN \left(\mathbf{0}, \mathrm{diag}(\sigma_{\ddot{x}}, \sigma_{\ddot{y}})\right)
\end{split}
\end{equation}
where $f$ defines the agent dynamics (an MDP can represent any general nonlinear dynamics function).

For the transition function $T$, when the action is a control action, the agent's next state is stochastically obtained from $f$. The agent-vehicle interactions switch the agent from not riding to riding ($\textsc{Board}$) and vice versa ($\textsc{Alight}$). They are deterministically successful (unsuccessful) if their pre-conditions are satisfied (not satisfied). For $\textsc{Board}$, the agent needs to be sufficiently close to a vehicle and sufficiently slow, while for $\textsc{Alight}$, it needs to currently be on a vehicle.

\textbf{Vehicles}: The transit routes are provided as streaming information from an external process (e.g. a traffic server). At each timestep the agent observes, for each currently active vehicle, its position $\bfp_t^i$ and estimated remaining route $\xi_t^i$ (waypoint locations and current ETA). New vehicles may be introduced at future timesteps arbitrarily. This online update scheme has been used previously to represent timetable delays~\cite{muller2009efficient}, though our model can also have speedups. For this work, we make two simplifying restrictions: (i) no rerouting: route waypoint locations, once decided, are fixed and traversed in order and (ii) bounded time deviations: estimates of waypoint arrival times increase or decrease in a bounded unknown manner between timesteps, i.e. $\lvert \tau_{j,t}^{i} - \tau_{j,t+1}^{i} \rvert \leq \Delta$.

\textbf{Objective}: Our single timestep reward function penalizes elapsed time and the agent's energy consumed (due to distance covered or hovering in place) if not riding on a vehicle:
\begin{equation}
\label{eq:reward}
R(s_{t},a,s_{t+1}) = -(\underbrace{\alpha (\lmbd \lVert \bfx_{t+1} - \bfx_{t} \rVert_2 + \lmbh \mathds{1}_{h})}_{\mathrm{energy}} +  \underbrace{(1-\alpha) 1}_{\mathrm{time}})
\end{equation}
where $\alpha \in [0,1]$ and $\mathds{1}_{h} = \mathds{1}[\lVert \dot{\bfx_t} \rVert < \epsilon]$ indicates hovering. The $\lmbd$ and $\lmbh$ coefficients encode the relative energy costs of distance and hovering, and the $1$ penalizes each timestep of unit duration (this formulation can represent the shortest distance and minimum time criteria). We vary $\alpha$ for experiments in~\cref{sec:experiments-results}. The agent must go from start state $\bfx_{s}$ to goal state $\bfx_{g}$.  The agent's performance criterion is the cumulative trajectory cost that \emph{depends on both discrete modes of transportation and continuous control actions}. It can ride on vehicles to save on distance, potentially at the cost of time and hovering.

DREAMR's underlying MDP is an online instance of \emph{Stochastic Shortest Paths} (SSP)~\cite{neu2010online}. An SSP is a special case of an undiscounted episodic MDP with only negative rewards i.e. costs, where the objective is to reach a terminal absorbing
state with minimum trajectory cost~\cite{bertsekas2005dynamic}. In online SSPs, there are two components. First, the identifiable system with known Markovian dynamics (our agent model). Second, an autonomous system which affects the
optimal decision strategy but is typically external (our transit vehicle routes).

\begin{algorithm}[t]
\caption{HHP for DREAMR}
\begin{algorithmic}[1]
\Procedure{Offline}{$\mathrm{CF,UF,K},\epsilon^{CF}$} \Comment{Sec.~\ref{sec:approach-closedloop}} \label{line:alg1-polst}
    \State $V^{CF},\pi^{CF} \gets \mathrm{PartialControlVI(CF,K,\epsilon^{CF})}$
    \State $V^{UF},\pi^{UF} \gets \mathrm{ApproximateVI(UF)}$
    \State $\pi^{Ri} \gets$ Deterministic Riding Policy
\EndProcedure \label{line:alg1-polend}
\Procedure{FlightAction}{$\bfx_t,\bfp_t,\tau_t,\pi^{UF,CF}$}
    \If {$\tau_t = \infty$} \Comment{Unconstrained Flight}
        \State \textbf{return} $\pi^{UF}(\{\bfx_t,\bfp_t\})$
    \Else \Comment{Constrained Flight}
        \State \textbf{return} $\pi^{CF}(\{\bfx_t,\bfp_t,\tau_t\})$
    \EndIf
\EndProcedure
\Procedure{RideAction}{$\bfx_t,\tau_t,\pi^{Ri}$}
    \State \textbf{return} $\pi^{Ri}(\{\bfx_t,\tau_t\})$
\EndProcedure
\Statex
\Procedure{Online}{$\bfx_s,\bfx_g,V^{CF,UF},\pi^{CF,UF},\pi^{Ri},\Delta T$}
    \State $t \gets 0, \ plan \gets true, \ lpt \gets 0$
    \State $G_t \gets (\{\bfx_s,\bfx_g\},\emptyset), \ mode \gets Flight$   
    \While{episode not terminated}
        \State Obtain current state $s_t$ \Comment{Eq.~\ref{eq:dreamr-state}}
        \State $G_t \gets \mathrm{UpdateVertices}(s_t,G_t)$ \Comment{No Edges} \label{line:alg1-gtupdate}
        \If {$plan = true$}
            \State $\zeta_{t} \gets \mathrm{ImplicitA^*}(G_t,V^{CF,UF})$ \Comment{Sec.~\ref{sec:approach-openloop}} \label{line:alg1-astar}
            \State $lpt \gets t, \ plan \gets false$
        \EndIf
        \State $\bfp_t,\tau_t \gets \mathrm{GetPositionTime}\big(\text{target}\left(\zeta_t[1]\right)\big)$
        \If {$mode = Flight$} \label{line:alg1-polact-st}
            \State $a_t \gets \mathrm{FLIGHTACTION}(\bfx_t,\bfp_t,\tau_t,\pi^{UF,CF})$
            \If {$a_t = \text{ABORT}$} \Comment{Sec.~\ref{sec:approach-interaction-aborting}} \label{line:alg1-abort}
                \State $plan \gets true$
            \EndIf
        \Else \Comment{$mode = Ride$}
            \State $a_t \gets \mathrm{RIDEACTION}(\bfx_t,\tau_t,\pi^{Ri})$
        \EndIf \label{line:alg1-polact-end}
        \State $s_{t+1},r_{t},mode \gets \text{Simulate}(s_t,a_t,mode)$ \label{line:alg1-sim}
        \If {$\bfx_{t+1} = \bfx_g$ }
            \State \textbf{return} success \Comment{Reached goal}
        \EndIf
        \State $t \gets t+1$
        \If {$t - lpt > \Delta T \ \text{or} \ mode \ \text{changed}$} \label{line:alg1-replan}
            \State $plan \gets true$ \Comment{Sec.~\ref{sec:approach-interaction-replanning}}
        \EndIf
    \EndWhile
    \State \textbf{return} failure
\EndProcedure
\end{algorithmic}
\label{alg:HHP-dreamr}
\end{algorithm}

\section{Approach}
\label{sec:approach}

We use the example of drones planning over car routes for exposition, but our formulation is general and is applicable to, say, terrain vehicles moving or riding on the beds of large trucks, or vehicles coordinating segments of drafting to reduce drag. \textit{No existing methods from multimodal routing or online SSPs can be easily extended to DREAMR}. The former do not consider real-time control under uncertainty while the latter are not designed for jointly optimizing discrete and continuous decisions. This motivates our novel framework.

For $n$ cars and $k$ waypoints per route, the full DREAMR state space has size $\mathbb{R}^{nk}$. It scales exponentially (for a continuous-valued base) with the numbers of cars and waypoints. Moreover, the dimensionality is significantly time-varying in a problem instance. Therefore, to solve DREAMR realistically, we must exploit its underlying structure:
\begin{enumerate}[label=(\roman*)]
  \item \emph{Decomposability}: Every route comprises three kinds of sub-routes: time-constrained flight to a car waypoint, riding on a car, and time-unconstrained flight. For each sub-route, \emph{at most one car is relevant}.
  \item \emph{Partial Controllability}: Optimal drone actions depend on the car positions and routes but cannot control them.
\end{enumerate}

Our Hierarchical Hybrid Planning (HHP) framework exploits DREAMR's structure with two \emph{layers}: a global open-loop layer with efficient graph search to plan a sequence of sub-routes, each of a particular mode (either \textit{Flight} or \textit{Ride}), and a local closed-loop layer executing actions in real-time using policies for the sub-routes. \emph{The open-loop layer handles online updates to transit options and produces discrete decisions while the closed-loop layer handles motion uncertainty and produces control actions}. Our framework balances deterministic replanning and stochastic planning.

\textbf{Our specific contributions} are the choices that unite techniques from multimodal routing and hierarchical stochastic planning --- implicit graph search, edges as macro-actions, value functions as edge weights, terminal cost shaping, replanning as switching between macro-actions (all described subsequently). The resulting HHP framework achieves both scalability and good quality solutions to DREAMR problems.~\Cref{alg:HHP-dreamr} summarizes the salient aspects of HHP.

\begin{figure}[t]
  \centering
  \fbox{\includegraphics[width=0.65\columnwidth]{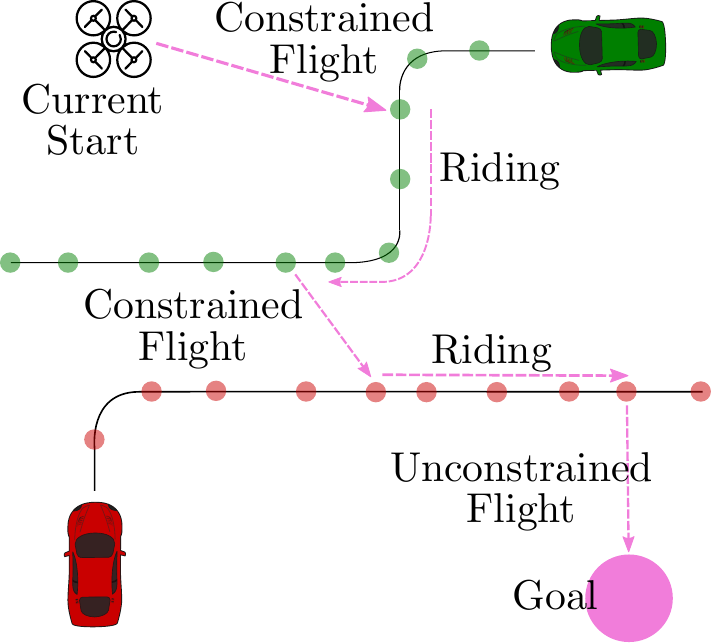}}
  \caption{The global open-loop layer computes the best sequence of sub-routes (edges) using a time-dependent DAG. There are three types of edges: constrained flight, riding, and unconstrained flight.}
  \label{fig:graph-layer}
\end{figure}

\subsection{Global Open-loop Layer}
\label{sec:approach-openloop}

\begin{figure*}[th]
  \centering
  \begin{subfigure}{0.66\columnwidth}
    \centering
    \includegraphics[width=0.92\textwidth, frame]{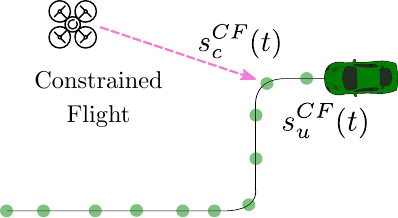}
    \caption{}
    \label{fig:interrupt-pre}
  \end{subfigure}
  \begin{subfigure}{0.66\columnwidth}
    \centering
    \includegraphics[width=0.92\textwidth, frame]{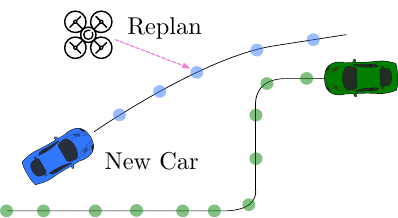}
    \caption{}
    \label{fig:interrupt-replan}
  \end{subfigure}
  \begin{subfigure}{0.66\columnwidth}
    \centering
    \includegraphics[width=0.92\textwidth, frame]{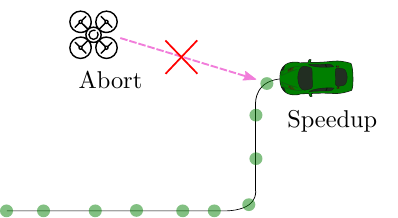}
    \caption{}
    \label{fig:interrupt-abort}
  \end{subfigure}
  \caption{(\subref{fig:interrupt-pre}) A constrained flight macro-action being executed can be interrupted in two ways: (\subref{fig:interrupt-replan}) replanning when a new car enters with a better path or the target car is delayed or (\subref{fig:interrupt-abort}) aborting when a successful \textsc{Board} becomes unlikely, say when a car speeds up suddenly.}
  \label{fig:interrupt}
\end{figure*}

\textbf{Graph Model}:
The prevalent framework for route planning is graph search~\cite{bast2016route}. We use a \emph{time-dependent}~\cite{pyrga2008efficient} directed acyclic graph (DAG) model with nonstationary edge weights. Every car waypoint is a vertex in the time-varying graph $G_t$. The $j\text{th}$ waypoint of the $i\text{th}$ car route is $v_{j,t}^{i} \equiv (\bfp_{j}^{i}, \tau_{j,t}^{i})$, with notation from~\cref{eq:dreamr-state}. The waypoint ETA $\tau_{j,t}^{i}$ is updated at each timestep. By the no rerouting and bounded deviation assumptions, once a route is generated, the waypoints are guaranteed to be traversed in order. Two more vertices $v_g \equiv (\bfx_{g}, \infty)$ and $v_{s} \equiv \left(\bfx_{s},t\right)$ represent drone goal and current drone source respectively.

The types of edges $(v \rightarrow v^{\prime})$ correspond to sub-routes:
\begin{itemize}
  \item
  Constrained Flight: $v^{\prime} = v_{j,t}^{i}$ (drone flies to a route waypoint and executes \textsc{Board}).
  \item
  Ride: $v = v_{j,t}^{i}\ , \ v^{\prime} = v^{i}_{j^{\prime},t^{\prime} > t}$ (drone executes \textsc{Alight} when the car reaches $v^{\prime}$).
  \item
  Unconstrained Flight: $v^{\prime} = v_g$ (drone flies to goal vertex).
\end{itemize}

Discretization is necessary to determine transfer points from a continuous route. The open-loop layer assumes that transfers can only happen at the pre-determined route waypoints. There is a tradeoff between solution quality and computation time as more transit points are considered per route, and the most suitable choice is a domain-specific decision.

\textbf{Route Planning}:
The open-loop layer runs $A^{*}$ search~\cite{hart1968formal} on $G_t$ (\cref{line:alg1-astar}) to obtain $\zeta_t$, the current best path (sequence of edges) from $v_{s}$ (current drone state) to $v_g$ (\cref{fig:graph-layer}). As the drone traverses the route and $G_t$ is updated (\cref{line:alg1-gtupdate}), it replans from scratch rather than incrementally repairing $G_t$, which is efficient only when few edges change~\cite{koenig2004incremental}. Our DAG is highly dynamic, with most edges updated as vertex timestamps change. Therefore, we use an \emph{implicit graph}, where edges are generated just-in-time by a successor function during the node expansion stage. Broadly speaking, the successor function enforces temporal constraints for CF edges and consistency for ride edges (the time-unconstrained goal $v_g$ is a potential neighbor for every expanded node). Implicit graphs have been used previously for efficient search in route planning~\cite{pajor2009multi,delling2009engineering}. For the $A^{*}$ heuristic, we use the minimum straight line traversal time while riding (therefore no energy cost),
\begin{equation}
\label{eq:heuristic}
 \mathrm{heuristic}(v,v_g) = (1 - \alpha) \cdot {\lVert v - v_g \rVert}/{\text{max car speed}}.
\end{equation}
The above heuristic is \emph{admissible}, assuming the maximum speed of a car is higher than that of a drone.

\textbf{Edge Weight}:
The edge weight function is a surrogate for~\cref{eq:reward}. For a \emph{ride edge}, the only cost is for elapsed time,
\begin{equation}
\label{eq:hh_edge_cost}
C(v_{j,t}^{i},v_{j^{\prime},t^{\prime}}^{i}) = (1 - \alpha) \cdot ( \tau_{j^{\prime},t^{\prime}}^{i} - \tau_{j,t}^{i}).
\end{equation}
However, from a flight edge, we get only Euclidean distance and the estimated time difference. The actual cost depends on the control actions of the policy while executing the edge. Therefore, we use the \emph{negative value function} of the macro-action policy, which is a better surrogate, as the corresponding edge weight; every edge encodes a state for the corresponding macro-action MDP (discussed subsequently). Consequently, we reduce average search complexity greatly by filtering out edges for infeasible or difficult connections.

\subsection{Local Closed-loop Layer}
\label{sec:approach-closedloop}

For the three kinds of edges, at most one car is relevant. This motivates state abstraction by treating edges as macro-actions with corresponding MDPs of smaller and fixed dimensionality~\cite{dietterich2000hierarchical,andre2002state}. A closed-loop policy for each MDP is obtained offline (\crefrange{line:alg1-polst}{line:alg1-polend}) and used by the local layer to execute the corresponding edge in real-time (\crefrange{line:alg1-polact-st}{line:alg1-polact-end}). Each macro-action policy only uses local information about the currently executing edge~\cite{hauskrecht1998hierarchical}.

\subsubsection{Constrained Flight (CF)}
The source is the drone vertex, $(\bfx_{t}, t)$, and the target is the connection point $(\bfp_{j}^{i}, \tau_{j,t}^{i})$ chosen by the global layer. Here, $\bfp_{j}^{i}$ is the position and $\tau_{j,t}^{i}$ the current ETA of the $i\text{th}$ car at its $j\text{th}$ waypoint.

The constrained flight (CF) macro-action policy $\pi^{CF}$ must control the drone from $\bfx_{t}$ to $\bfp_{j}^{i}$ and slow down to a hover \emph{before the car arrives} to execute \textsc{Board} successfully. It also needs to adapt to changes in $\tau_{j,t}^{i}$. The CF problem is \emph{partially controlled} as the drone action does not affect the car's ETA. Therefore, we build upon a partially controllable finite-horizon MDP approach originally applied to collision avoidance~\cite{kochenderfer2011collision}. It decomposes the problem into controlled and uncontrolled subproblems, solved separately offline and combined online to obtain the approximately optimal action (\cite{bertsekas2005dynamic}, Sec. 1.4).

For CF, the state is $s^{CF}(t) = (s_{c}^{CF}(t), s_{u}^{CF}(t))$ where
\begin{equation}
    \begin{aligned}
      &s_{c}^{CF}(t) = (\bfx_t - \bfp_{j}^{i}, \dot{\bfx}_{t}) \ &\text{Controlled}\\
      &s_{u}^{CF}(t) = (\tau_{j,t}^{i} - t) \ &\text{Uncontrolled}
    \end{aligned}
\end{equation}
represents relative drone state and time to car ETA respectively. The MDP episode terminates when $s^{CF}_{u}(t) < \epsilon^{CF}$, where $\epsilon^{CF}$ is a small time threshold for the car ETA.

The action space is for drone control, i.e. $\calA^{CF} \equiv \mathcal{U}$. The transition function $T^{CF}$ uses drone dynamics from~\cref{eq:drone-dynamics}. For non-terminal states, the reward function $R^{CF}$ is the same as~\cref{eq:reward}. A \emph{constrained flight edge encodes a state} of the CF MDP, so the edge weight for the global graph search is set to the negative value of the encoded state. We use terminal cost shaping (discussed subsequently) such that the value function approximates the drone dynamics cost.

The partial control method computes a horizon-dependent action-value function for the controlled subproblem ($Q_{0:K}$ and $Q_{\bar{K}}$ for out-of-horizon, denoted with $Q_{0:K,\bar{K}}$), assuming a distribution over episode termination time. The uncontrolled subproblem dynamics define probability distributions over termination time for the uncontrolled states ($D_{0:K}$ and~$D_{\bar{K}} = 1 - \sum_{k=0}^{K} D_k$), where $D_k(s_u)$ is the probability that $s_u$ terminates in $k$ steps.  The $Q_k$ and $D_k$ values are computed offline and used together online to determine the (approximately) optimal action for the state $(s_c,s_u)$,
\begin{equation}
\label{eq:pcmdp-action}
a^{*} = \underset{a}{\mathrm{argmax}} \ D_{\bar{K}}(s_u) Q_{\bar{K}}(s_c,a) + \sum_{k=0}^{K} D_k(s_u) Q_k(s_c,a).
\end{equation}
The partial control decomposition stores $Q_K^{*}$ compactly in $O(K|A||S_c| + K|S_u|)$ over $O(|K||A||S_c||S_u|)$~\cite{kochenderfer2011collision}.

We compute $Q^{CF}_{0:K,\bar{K}}$ (and value function $V^{CF}$ and policy $\pi^{CF}$) offline. Our uncontrolled state $s^{CF}_{u}(t) = (\tau_{j,t}^{i} - t)$ is just the time to the car's ETA, i.e. the CF termination time.
If $\tau_{j,t}^{i}$ is provided as a distribution over ETA, we obtain $D_{0:K,\bar{K}}(s^{CF}_u)$ directly from $(\tau_{j,t}^{i} - t)$. Otherwise, if it is a scalar estimate $\hat{\tau}_{j,t}^{i}$, we construct a normal distribution $\calN(\hat{\tau}_{j,t}^{i} - t, \sigma_{j,t}^{i})$, where $\sigma_{j,t}^{i}$ is the standard deviation of the observed ETAs so far, and sample from it to compute $D_{0:K,\bar{K}}(s^{CF}_u)$. 

\textbf{Terminal Cost Shaping}:
A CF episode terminates when $s^{CF}_{u}(t) < \epsilon^{CF}$, i.e. at horizon $0$. Let $\chi^{CF}_0$ be the set of successful terminal states, where relative position $(\bfx - \bfp_{j}^{i})$ and velocity $\dot{\bfx}$ are close enough to $0$ to \textsc{Board}. Typically, successful terminal states for a finite-horizon MDP are given some positive reward to encourage reaching them. However, there is no positive reward in~\cref{eq:reward} (recall DREAMR is an SSP). Therefore, rewarding a hop will make $V^{CF}$ an \emph{unsuitable surrogate of the true cost} and a poor edge weight function; the open-loop layer will prefer constrained flight edges likely to succeed whether or not they advance the drone to the goal. Instead, we have a sufficiently \emph{high cost penalty for unsuccessful terminal states} to avoid choosing risky CF edges. Any terminal state $s^{CF}_{c}(0)$ not in $\chi^{CF}_0$ is penalized with $\phi^{CF}$ lower bounded by the maximum difference between any two $K$-length control sequences $\bfu^{i}_{1:K}$ and $\bfu^{j}_{1:K}$,
\begin{multline}
\label{eq:failure_cost}
  \phi^{CF} > \underset{\bfu^{i,j}_{1:K}}{\max} \sum_{k=1}^{K} \bigl( R^{CF}(s^{CF}_c(k),\bfu_{k}^{i}) \\
  - R^{CF}(s^{CF}_c(k),\bfu_{k}^{j}) \bigr),
\end{multline}
such that any high-cost successful trajectory is preferred to a lower-cost unsuccessful one. The $\phi^{CF}$ value can be computed in one pass offline. This \emph{terminal pseudo-reward} makes $\pi^{CF}$ reusable for all CF edges (c.f~\cite{dietterich2000hierarchical}, pages \numrange{248}{250}).

\subsubsection{Riding and Unconstrained Flight}
We discuss these simpler macro-actions briefly. A riding edge is between two waypoints, after the successful execution of \textsc{Board} and up to the execution of \textsc{Alight}. This is an \emph{uncontrolled problem} where the (deterministic) policy is: do nothing until the car reaches the waypoint and then \textsc{Alight}. An unconstrained flight (UF) edge represents controlling the drone from the current waypoint to the goal, which is a time-unconstrained drone state. Therefore, we use an infinite horizon MDP, and run value iteration until convergence to obtain the control policy $\pi^{UF}$ for unconstrained flight (we could also have used an open-loop controller here). The value function $V^{UF}$ for the unconstrained flight MDP is used for the edge weight.

For both CF and UF, we discretized the drone state space with a multilinear grid evenly spaced along a cubic scale, and used interpolation for local value function approximation. The cubic scale has variable resolution discretization~\cite{munos1999variable} and \emph{better approximates the value function} near the origin, where it is more sensitive. The discretization limits were based on the velocity and acceleration bounds.

\begin{figure}[t]
    \centering
    \includegraphics[width=0.7\columnwidth]{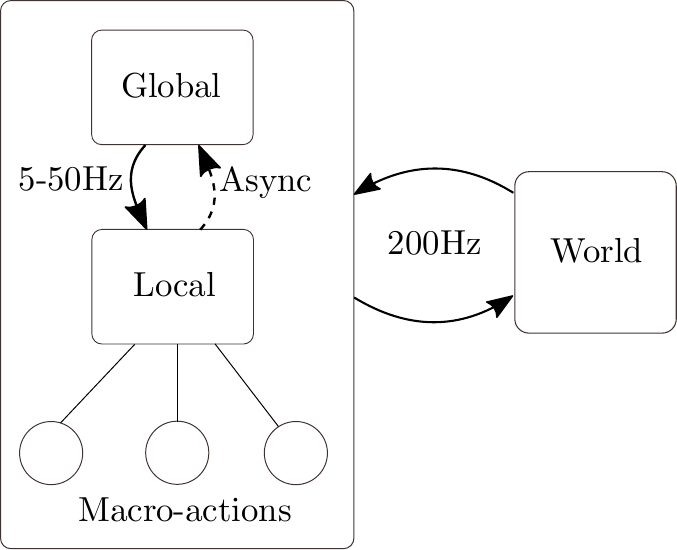}
    \caption{On our problem scenarios (\cref{sec:experiments}), the HHP framework executes actions at \SI{200}{\hertz}, through the local closed-loop layer (fixed-size MDPs so time is agnostic to route network size). The global layer plans routes on the graph at a frequency of \SIrange{5}{50}{\hertz}, depending on network size, which determines the local layer policy to invoke. The global layer bottlenecks the HHP framework only when macro-actions terminate asynchronously.}
    \label{fig:archi}
\end{figure}

\subsection{Interleaving Planning and Execution Layers}
\label{sec:approach-interaction}

The car route updates require significant adaptivity. New cars may improve options, a delay may require \textsc{Alight}, and a speedup may invalidate the CF edge. The global open-loop layer keeps replanning as the car routes are updated, to obtain the current best
path $\zeta_t$. The local closed-loop layer is (somewhat) robust to uncertainty within the macro-action but is unaware of the global plan, and may need interrupting so a new $\zeta_t$ can be executed.~\Cref{fig:interrupt} depicts scenarios that motivate interleaved planning and execution.

\subsubsection{Open-Loop Replanning}
\label{sec:approach-interaction-replanning}

HHP uses periodic and event-driven replanning, which effectively obtains high-quality schedules~\cite{church1992analysis}. The period $\Delta T$ is the ratio of layer frequencies and the events are aborting constrained flight or mode switches (\cref{line:alg1-abort,line:alg1-replan}).
If the first edge of the new path is different than the current one, the correspondingly different macro-action policy is invoked.
\emph{This switching between macro-actions is seamlessly achieved by switching between paths in our framework} since edge weights are based on macro-action value functions and edges are equivalent to macro-action states.
For hierarchical stochastic planning, switching between macro-actions rather than committing to one until it terminates guarantees improvement (c.f.~\cite{sutton1999improved} for proof).

\subsubsection{Aborting Constrained Flight}
\label{sec:approach-interaction-aborting}

Aborting the CF macro-action is useful when the target car speeds up and the drone might miss the connection. For $k = 1\ldots K$, we compute (offline) the worst value of any controlled state (least likely to succeed from horizon $k$),  $\underaccent{\bar}{V}^{CF}(k) = \min_{s^{CF}_c,a} Q^{CF}_{k}(s^{CF}_{c},a)$.  During macro-action execution (online),
\begin{equation}
\label{eq:beta-abort}
V^{CF}_k(s^{CF}_c) < \beta \cdot \underaccent{\bar}{V}^{CF}(k) \implies \ \text{Abort CF}
\end{equation}
where $\beta \in [0,1]$ is a \emph{risk factor}. The graph search successor function also uses~\cref{eq:beta-abort} to reject edges that are unlikely to yield successful connections. The closer $\beta$ is set to 1, the greater the risk we take for the connection. For $\beta = 1$, the CF macro-action never aborts (but may still be interrupted by the global layer replanning). By incorporating abort into $\pi^{CF}$, we can reason about it at the higher closed-loop frequency. We evaluate the effect of $\beta$ in~\cref{fig:res-tradeoff}, but the framework could adaptively use multiple $\beta$-policies at the same time.

\section{Experiments}
\label{sec:experiments}

We used Julia and the POMDPs.jl framework~\cite{egorov2017pomdps}.~\Cref{fig:archi} shows a simplified architecture.
Simulations were done with \SI{16}{\gibi\byte} RAM and a $6$-core \SI{3.7}{\giga\hertz} CPU.~\Cref{table:1} summarizes HHP's scalability to network size through the open-loop query time, which is competitive with state-of-the-art multimodal journey planning~\cite{giannakopoulou2018multimodal} (c.f. Table 7 in their Appendix).
A perfect comparison is not feasible as they have different technical specifications and use a public transit network with explicit edges, while we use an implicit dynamic graph. Still, their Berlin graph ($|E| = \num{4.3e6}$, c.f. their table 2) is comparable in size to row 2 of our~\cref{table:1}. Their \texttt{MDTM-QH} algorithm with Bus and Train (2 modes like ours) without the ALT preprocessing (inapplicable for dynamic graphs) takes
$\SI{14.97}{\milli\second}$ per query while ours takes $\SI{8.19}{\milli\second}$ for the most expensive first search. The average computation time for our local layer is $\SI{5}{\milli\second}$ (independent of network size).

\begin{figure}[t]
  \centering
  \fbox{\includegraphics[width=0.6\columnwidth]{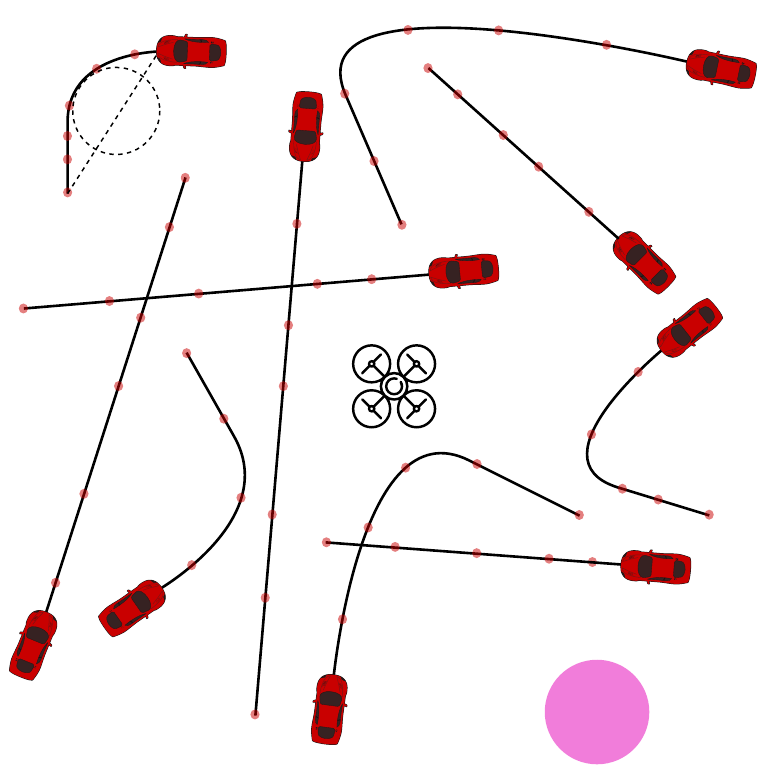}}
  \caption{A simplified visualization of the first epoch in an episode. The drone starts at the center and the goal location is near a corner. The number of cars is much smaller than for the actual problems, for better viewing. The route on the top left shows how an L-shaped path is synthesized.}
  \label{fig:dset}
\end{figure}

\subsection{Baseline}
\label{sec:experiments-baseline}

To baseline our HHP framework, we used receding horizon control (RHC), which repeatedly does open-loop planning. The baseline uses HHP's graph search to choose routes, with a nominal edge weight for flight distance and time. Flight edges are traversed by repeated non-linear trajectory optimization:
\begin{equation}
\label{eq:rhc}
\begin{aligned}
& \underset{\bfu_{0:N-1}}{\mathrm{argmin}}
& & \sum\limits_{k=0}^{N-1} \alpha (\lmbd \lVert \bfx_{k+1} - \bfx_{k} \rVert + \lmbh \mathds{1}[\dot{\bfx}_{k} < \epsilon]) \\
& & & \ \ \ \ \ \ \ + (1-\alpha) \mathds{1}[\bfx_{k+1} \notin \chi_F]  \\
& \text{subject to}
& & \bfx_{k+1} = f(\bfx_k,\bfu_k), \ \bfu_{0:N-1} \in U \\
& & & \bfx_0 = \bfx_t, \ \bfx_{N} \in \chi_{F}
\end{aligned}
\end{equation}
with shared symbols from~\cref{eq:drone-dynamics} and~\cref{eq:reward}. The RHC horizon $N$ is adjusted depending on car ETA for CF, and fixed for UF, and $\chi_F$ is the set of successful terminal states (e.g. $\chi_{F} = \chi^{CF}_0$ for constrained flight).  We attempt \textsc{Board} at the waypoint if the pre-conditions are satisfied, or else we replan. Ride edges use the same strategy as in HHP.

\subsection{Problem Scenarios}
\label{sec:experiments-dataset}

We are evaluating a higher-level decision-making framework so we abstract away physical aspects of obstacles, collisions, and drone landing. We also mention only illustrative experimental parameters in the interest of space. The spatiotemporal scale is chosen to reflect real-world settings of interest. We simulate routes on a grid representing \SI{10}{\kilo\metre} $\times$ \SI{10}{\kilo\metre} (approximately the size of north San Francisco). We have scaled velocity and acceleration limits for the drone. Each episode lasts for $30$ minutes (\SI{1800}{\second}), with $360$ timesteps or epochs of \SI{5}{\second} each. An episode starts with between $50$ to $500$ cars (unlike the fixed-size settings for scalability analysis in~\cref{table:1}) and more are added randomly at each subsequent epoch (up to twice the initial number). 

\begin{figure}[t]
    \centering
    \includegraphics[width=0.95\columnwidth]{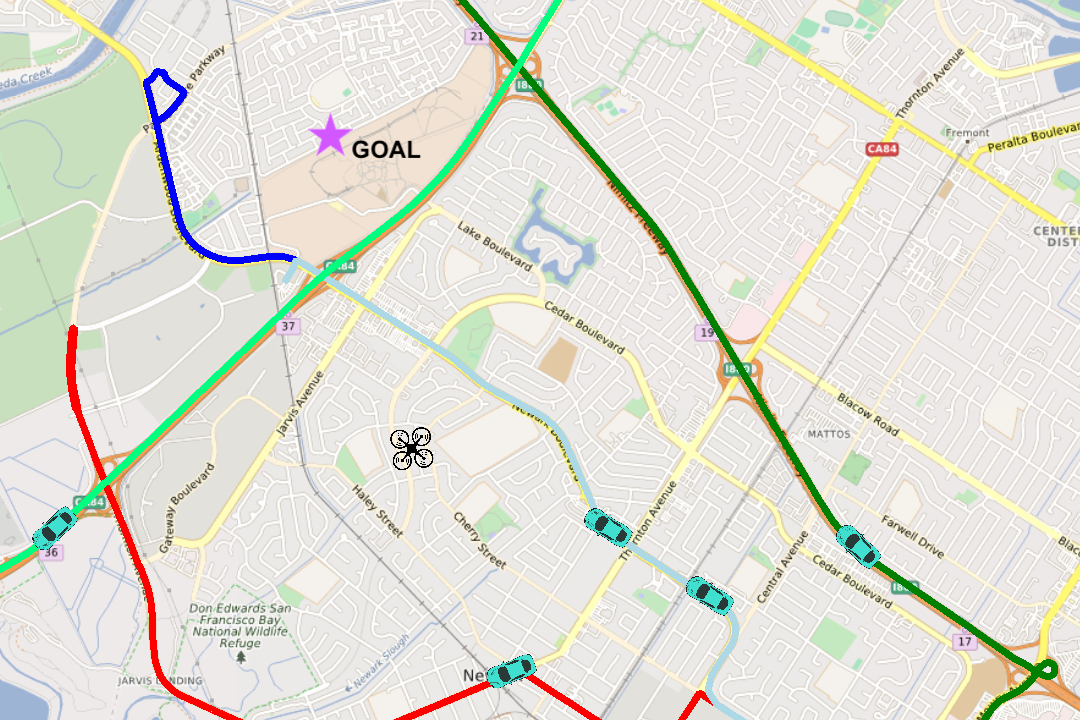}
    \caption{The problem scenario above is a special case of our experimental setup, with only the distance scale parameters requiring modification. Car routes on the road network are represented as a sequence of time-stamped GPS waypoints. We have shown only a few cars for illustration.}
    \label{fig:maps}
\end{figure}

A new car route is generated by first choosing two endpoints more than $\SI{2}{\kilo\metre}$ apart. We choose a number of route waypoints from a uniform distribution of $5$ to $15$, denoted $\mathrm{U}(5,15)$, and a route duration from $\mathrm{U}(100,900)$s. Each route is either a straight line or an L-shaped curve. The waypoints are placed along the route with an initial ETA based on an average car speed of up to \SI{50}{\metre\per\second}. See~\cref{fig:dset} for a simplified illustration.

The scheme for updating route schedules is based on previous work~\cite{muller2009efficient} but is simulated to be more adversarial. At each epoch, the car position is propagated along the route and the ETA at each remaining waypoint is perturbed with $p = 0.75$ within $\pm$\SI{5}{\second} (highly dynamic). The route geometry is simple for implementation, but we represent routes as a sequence of waypoints. This is \emph{agnostic to route geometry or topology} and can accommodate structured and unstructured scenarios. For instance, the urban street-map scenario shown in~\cref{fig:maps} can be seamlessly represented by our setup.

In addition to transit routes, we have a simulator for the agent dynamics model (we use a double integrator) and agent-transportation interactions (\cref{line:alg1-sim}). The updated route information at each epoch is provided with the current MDP state, formatted as in~\cref{eq:dreamr-state}. The drone starts at the grid center, while the goal is near a corner. The code repository\footnotemark has a video for two scenarios that illustrate HHP's behavior.
\footnotetext{The Julia code is at \url{https://github.com/sisl/DreamrHHP}}

\subsection{Results}
\label{sec:experiments-results}

\begin{table}[t]
\centering
\begin{tabular}{@{} rrrrr @{}}
    \toprule
    $\lvert V \rvert$ & $\lvert V \rvert^2$ & Setup & $1^{\mathrm{st}}$ search & $25\%$ searches\\
    \midrule
    $1000$         & \num{e6}   & $0.98$ & $1.81$   & $\SIrange[range-phrase = -]{0.3}{0.9}{}$\\
    $2000$         & \num{4e6}    & $2.02$ & $8.19$   & $\SIrange[range-phrase = -]{1}{5}{}$\\
    $5000$         & \num{2.5e7}  & $4.97$ & $175$    & $\SIrange[range-phrase = -]{5}{25}{}$\\
    \num{e4} & \num{e8}    & $9.61$ & $610$    & $\SIrange[range-phrase = -]{30}{75}{}$\\
    \bottomrule
\end{tabular}
\caption{The running time of HHP's graph search layer varies with problem size. All times are in \SI{}{\milli\second} and averaged over 100 runs in each case. The number of vertices $|V|$ is the total number of route waypoints and is controlled by fixing the number of cars and the number of route waypoints per car (\cref{sec:experiments-dataset}).
For implicit graphs, with edges generated on-the-fly, $|E|$ is not applicable but the worst case search complexity is $O(|V|^2)$.
The setup time scales linearly with $|V|$ as expected, the first search scales somewhat inconsistently due to memory handling, and the subsequent replanning is particularly efficient (a range of search times is shown). After the first $25\%$ of searches, many car routes terminate, after which further searches are trivially fast (\SI{0.001}{\milli\second}) and not informative of scalability.}
\label{table:1}
\end{table}

\begin{figure}[t]
    \includegraphics[width=\columnwidth]{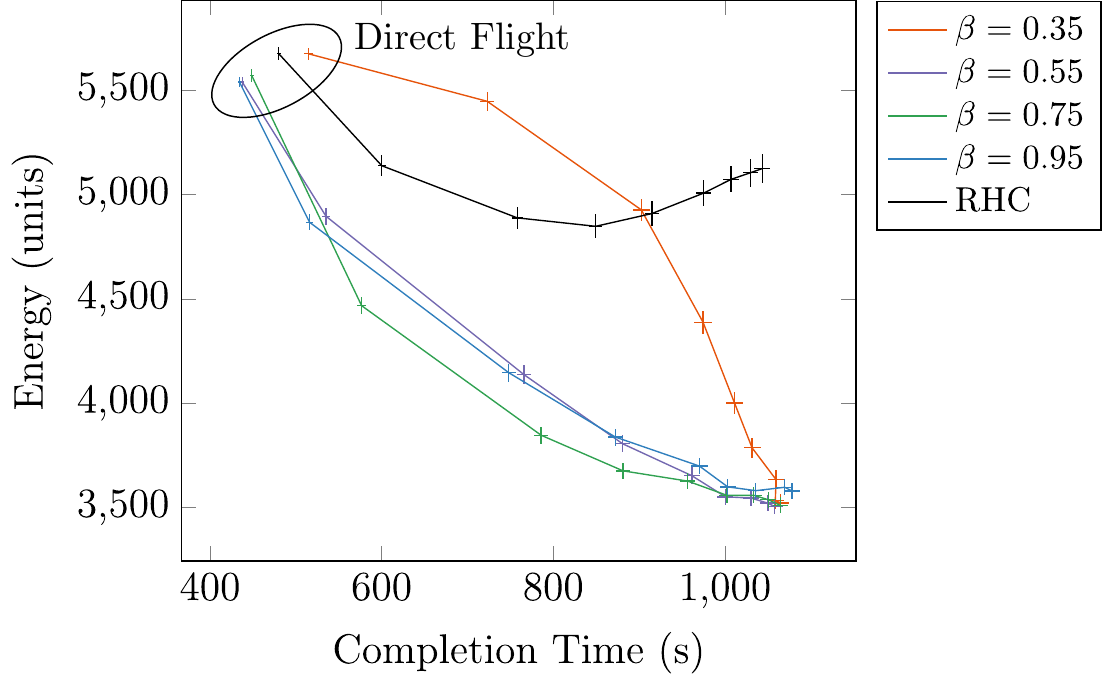}
    \caption{Our approach (HHP) with varying $\beta$ risk parameters is compared against receding horizon control (RHC), with minimizing energy increasingly prioritized over time from left to right. The 9 points on each curve show the energy-time tradeoff for 9 linearly spaced values of $\alpha$ (\cref{eq:reward}) from 0 (left) to 1 (right), averaged over 1000 different episodes in each case. The error bars depict the standard error of the mean for each metric. The tradeoff curves for HHP are generally significantly better and more consistent than for RHC, except for $\beta=0.35$.}
    \label{fig:res-tradeoff}
\end{figure}

We compared RHC against four variants of HHP using different $\beta$ values (the risk parameter in~\cref{sec:approach-interaction}). We used 9 values of the $\alpha$ energy-time tradeoff parameter from~\cref{eq:reward}, linearly spaced from $0$ to $1$. For each corresponding $R$, we evaluated the two approaches over 1000 simulated episodes (different from~\cref{table:1} episodes) and plotted in~\cref{fig:res-tradeoff} their average performance with the two components of $R$, the energy consumed and time to reach the goal.

For $\alpha = 0$ (left-most), the objective is to minimize time by \emph{typically} flying directly to goal at full speed without riding. Subsequent points (increasing $\alpha$) correspond to prioritizing lower energy consumption by more riding, typically taking more time, although sometimes riding may save both time and energy. \textbf{HHP significantly dominates RHC on both metrics}, with RHC requiring up to $40\%$ more energy. Compared to HHP, flying directly to the goal (the top-left-most point) requires up to $60\%$ more energy.

\begin{figure}[t]
    \centering
    \includegraphics[width=0.65\columnwidth]{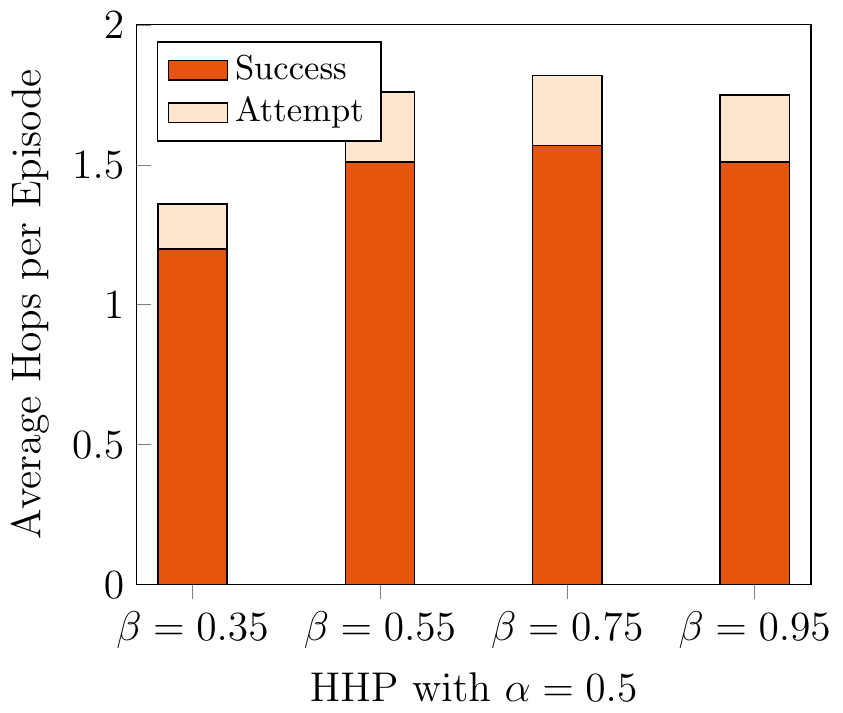}
    \caption{For low $\beta$ (0.35), HHP is conservative and attempts only easy CF edges, with a higher success rate. As $\beta$ increases, HHP optimistically attempts hops but fails slightly more often. There is a slight dip in attempts for the highest $\beta$ (0.95) because some unlikely constrained flight edges are abandoned as infeasible midway by the global layer, before being attempted.}
    \label{fig:hops-beta}
\end{figure}

A higher $\beta$ for $\pi^{CF}$ makes HHP optimistically attempt constrained flight (CF) connections. Lower values are more sensitive; $\beta= 0.35$ has the poorest energy-time tradeoff due to being risk-averse to potentially useful connections.~\Cref{fig:hops-beta} further analyzes the effect of $\beta$ on CF connections. Increased optimism can affect success rates for the attempts.

The difference between HHP and RHC varies with $\alpha$. For $\alpha=0$ they both fly straight to goal and any small differences are due to the underlying control method. However, the performance gap widens with increasing $\alpha$ as more riding is considered to save energy. For CF edges, RHC's nominal edge weight \emph{only accounts for time difference and flight distance}, but not hovering or control action costs. The negative value function of the encoded state better estimates the traversal cost and penalizes the CF edge if failure is likely. HHP is thus less likely to commit to difficult connections (depending on $\beta$). Furthermore, \textbf{RHC's underestimation of the true energy cost is exacerbated as energy consumption is prioritized}, which accounts for the greater deviation in RHC's tradeoff curve as $\alpha$ approaches 1 on the right.

We provide an illustrative example about the edge weights. Consider two CF edges for which HHP succeeds on the first but aborts on the second. For the success, the nominal edge weight was $3233$ and the value function cost was $3644$; the actual execution trace until \textsc{Board} incurred $3723$ units of cost. In the abort case, the nominal edge weight estimate was $5418$ and the value function estimate was $6117$. For edges likely to succeed, HHP's cost estimate is more accurate, while for likely failures, HHP's cost estimate is higher and the edges less preferred. This example is anecdotal but this effect cumulatively contributes to the performance gap.

\section{Conclusion}
\label{sec:conclusion}

We introduced Dynamic Real-time Multimodal Routing (DREAMR) and formulated it as an online discrete-continuous SSP. Our hierarchical hybrid planning framework exploits DREAMR's decomposability, with a global open-loop layer using efficient graph-based planning for discrete decisions and a local closed-loop policy layer executing continuous actions in real-time. Our framework scales to large problems, and interleaves planning and execution in a principled manner by switching macro-actions. Empirically, we are competitive with state-of-the-art multimodal routing for high-level plans. We also greatly outperform a receding horizon control baseline for the energy-time tradeoff.

\noindent
\textbf{Limitations and Future Work}:
This paper summarizes initial work introducing DREAMR and designing an HHP framework. There are several limitations motivating future research. The algorithmic choices we made for our framework work well but a comparative study with other components would be valuable. Our design and analysis is empirical; theoretical analysis of optimality for continuous-space online stochastic sequential problems requires several restrictive modeling assumptions. There is a taxonomy of DREAMR problems, depending on single or multiple controllable agents, control over transportation, and performance criteria. Solving them will require extensions and modifications to our current framework. The underlying MDP model will be explored in the context of other hybrid stochastic decision making problems. The grid scenarios are sufficient to evaluate scalability and decision-making, but we will create city-scale datasets as well. On the engineering side, a more sophisticated simulator is also of interest.

{\footnotesize

}

\end{document}